\documentclass{article}
\usepackage{spconf,amsmath,graphicx}
\usepackage{indentfirst}

\usepackage{algorithm}
\usepackage{amsmath}
\usepackage[noend]{algpseudocode}
\usepackage{setspace}  
\title{\vspace{-1.5em}A supervised STDP-based Training Algorithm for Living Neural Networks \vspace{-0.5em}}
\name{Yuan Zeng$^*$ \quad Kevin Devincentis$^\dag$  \quad Yao Xiao$^\ddag$ \quad Zubayer Ibne Ferdous$^*$ }
{Xiaochen Guo$^*$ \quad  Zhiyuan Yan$^*$ \quad Yevgeny Berdichevsky$^*$$^\S$   \vspace{-0.5em}}
\address{Lehigh University, $^*$Electrical and Computer Engineering Department $^\S$Bioengineering Department \\ 
$^\dag$Carnegie Mellon University, Electrical and Computer Engineering Department \\
$^\ddag$University of Science and Technology of China, School of Gifted Young 
\vspace{-1em} \thanks{This work is supported in part by a CORE grant from Lehigh University and in part by the National Science Foundation under Grant ECCS-1509674.}}

\begin{document}
\maketitle
\vspace{-1em}
\begin{abstract} 
\vspace{-0.6em}
\ \; \;\;\;Neural networks have shown great potential in many applications like speech recognition, drug discovery, image classification, and object detection. Neural network models are inspired by biological neural networks, but they are optimized to perform machine learning tasks on digital computers. The proposed work explores the possibility of using living neural networks \textit{in vitro} as the basic computational elements for machine learning applications. A new supervised STDP-based learning algorithm is proposed in this work, which considers neuron engineering constraints. A 74.7\% accuracy is achieved on the MNIST benchmark for handwritten digit recognition.
\end{abstract}
\vspace{-0.5em}


\vspace{-1.5em}
\section{Introduction and Motivation}
\label{sec:intro}
\vspace{-1em}
Artificial Neural Network (ANN) and Spiking Neural Network (SNN) are two brain inspired computational models, which have shown promising capabilities for solving problems such as face detection \cite{faceDetection1} and image classification \cite{faceDetection2}. Computer programs based on neural networks can defeat professional players in the board game Go \cite{go} \cite{gozero}. \!ANN relies on numerical abstractions to represent both the states of the neurons and the connections among them, whereas SNN uses spike trains to represent inputs and outputs and mimics computations performed by neurons and synapses \cite{ANNSNN}. Both ANN and SNN are models extracted from biological neuron behaviors and optimized to perform machine learning tasks on digital computers. 

In contrast, the proposed work explores whether biological living neurons \textit{in vitro} can be directly used as basic computational elements to perform machine learning tasks.  Living neurons can perform ``computation'' naturally by transferring spike information through synapses. The energy consumption is 100000$\times$ more efficient than hardware evaluation \cite{brainEnergy}. Living neurons have small sizes (4 to 100 micrometers diameter) \cite{size}, and adapt to changes. 

While precise control of living neural networks is challenging, recent advances in optogenetics, genetically encoded neural activity indicators, and cell-level micropatterning open up possibilities in this area \cite{Microfluidics1} \cite{Microfluidics2} \cite{electrodeArrays}. Optogenetics can label individual neurons with different types of optically controlled channels, which equip \textit{in vitro} neural networks with optical interfaces. Patterned optical stimulation and high-speed optical detection allow simultaneous access to thousands of \textit{in vitro} neurons. In addition, the invention of micropattern \cite{micropattern} enables modularized system design.

To the best of our knowledge, this work is the first to explore the possibility of using living neurons for machine learning applications. By considering neuron engineering design constraints, a new algorithm is proposed for easy training in future biological experiments. A fully connected spiking neural network is evaluated on MNIST dataset using NEURON simulator \cite{neuron}. A 74.7\% accuracy is obtained based on a biologically-plausible SNN model, which is a promising result that demonstrates the feasibility of using living neuron networks to compute. 
\vspace{-1.2em}
\section{Methods}
\label{sec:algorithm}
\vspace{-1em}
Spiking neural network is a model that closely represents biological neuron behavious. In biological neural network, neurons are connected through plastic synapses. A spike of a pre-synaptic neuron will change the membrane potential of a post-synaptic neuron. The impact of the spikes at different time from all of the pre-synaptic neurons will be accumulated at the post-synaptic neuron. Post synaptic spikes are generated when the membrane voltage of a neuron exceeds a certain threshold. In SNN, information is represented as a series of spikes, and the accumulative effect of the pre-synaptic spikes are also modeled.

This work models biological neural networks using the Hodgkin-Huxley (HH) neuron model \cite{HH} and spike-based data representation. One major difference between the proposed work and prior SNN based models \cite{unsupervised1} \cite{unsupervised2} \cite{unsupervised3} \cite{supervised1} \cite{supervised2} \cite{supervised3} is that this work aims to explore the potential of using biological living neurons as the functional devices, while prior works focused on the computational capability of the neuron model. Neuron engineering design constraints lead to different design choice for input encoding, network topology, neuron model, learning rule, and model parameters.

\vspace{-1.2em}
\subsection{Network topology}
\vspace{-0.6em}

Neural connectivity in human brain is complex and has different types of topologies in different parts of the nervous system. To understand the network functionality, a simple network topology is built, where all input neurons are connected to all output neurons through synapses (Fig. 1).

Due to bioengineering constraints, images from the MNIST dataset, which have 28 $\times$ 28 pixels, are compressed to 14 $\times$ 14 pixels. As a result of this simplification, the network has 196 input neurons, each corresponding to one pixel. Only black pixels generate spikes, and all input spikes occur simultaneously.

Output for the network is a vector of spiking state for output neurons, ``1'' represents a spike, ``0'' means no spike. Each output neuron is associated with a group index from 0 to 9, which can be defined artificially. The index of the group with the largest number of spiking neurons will be considered as the network output. In our experiment, 300 output neurons are used. Every 30 consecutive neurons belong to one group, for example, the first 30 belong to group 0. 

\vspace{-1em}
\begin{figure}[htb]
\begin{minipage}[b]{1.0\linewidth}
  \centering
  \centerline{\includegraphics[width=8cm]{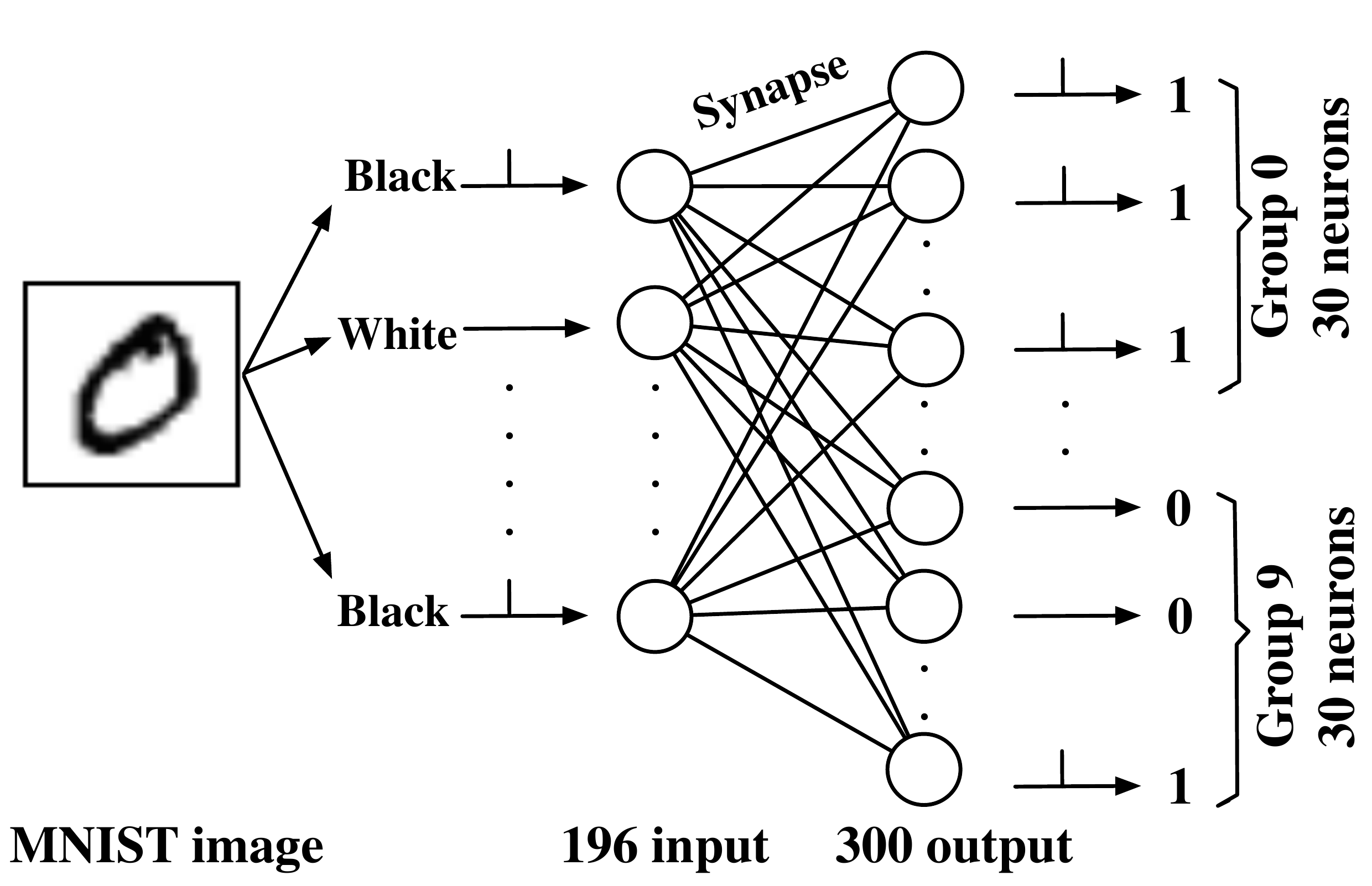}}
\end{minipage}
\vspace{-2em}
\caption{Network topology.}
\label{fig:stdp}
\end{figure}
\vspace{-1em}

\vspace{-1em}
\subsection{Neuron model}
\vspace{-0.5em}

To capture the realistic neuron dynamics, the Hodgkin-Huxley (HH) model \cite{HH} is used in the simulation, which models the electro-chemical information transmission of biological neurons with electrical circuit. HH model has been successfully verified by numerous biological experimental data and it is more biologically accurate than other simplified model such as Integrate and Fire model \cite{if}.

A spike will be generated if the membrane potential for a neuron exceeds a certain threshold. \!However, after a strong current pulse excites a spike, \!there will be a period during which current pulse at the same amplitude cannot generate another spike, \!which is referred to as the refractory period \cite{refractory}. 

\vspace{-1em}
\subsection{Learning rule}
\vspace{-0.5em}

The plasticity of synapses between neurons is important for learning. Connection strength changes based on precise timing between pre- and post-synaptic spikes. This phenomenon is called Spike Timing Dependent Plasticity (STDP) \cite{STDP}.

\begin{figure*}[htb]
\vspace{-1.5em}
\begin{minipage}[b]{1.0\linewidth}
  \centering
  \centerline{\includegraphics[width=17cm]{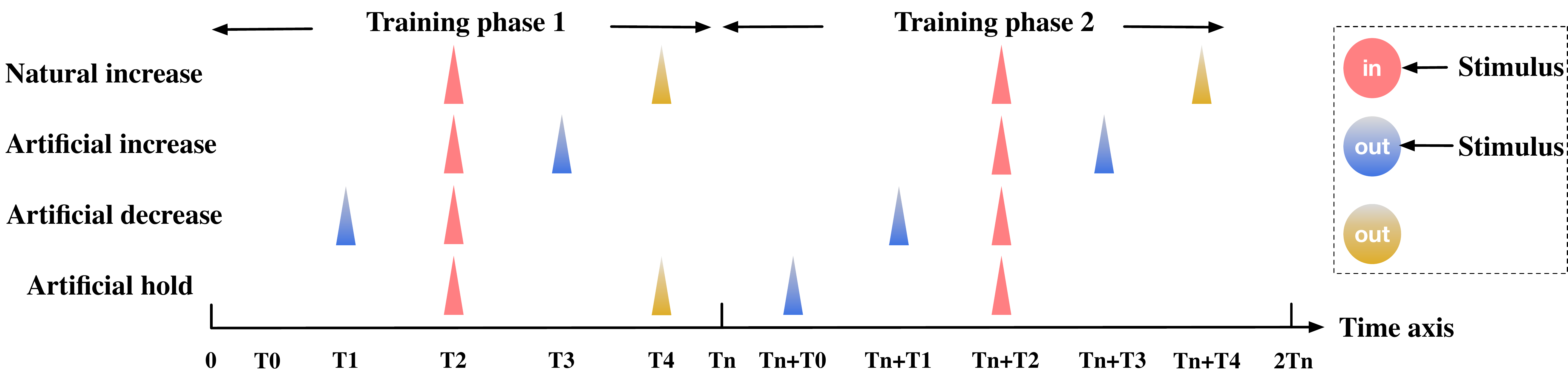}}
\end{minipage}
\vspace{-2.2em}
\caption{Four mechanisms in the supervised STDP training.}
\label{fig:stdp}
\vspace{-1.5em}
\end{figure*}

Eqs. (1)-(4) describe the STDP rule \cite{nmodl} used in this work. In this rule, the weight changes are proportional to spike trace \cite{trace}. $t_{pre}$ and $t_{post}$ denote the time a pre- and post-synaptic, respectively, spikes arrive; $t_{pre}\prime$ and $t_{post}\prime$ represent the arrival time of previous pre- and post-synaptic spikes respectively. $ALTP$ and $ALTD$ are the amplitudes of trace updating for potentiation and depression respectively; and $aLTP$ and $aLTD$ are the potentiation and depression learning rate respectively. Each pre-synaptic spike arrival will update the pre-synaptic trace $P$ according to  Eq. (1), post-synaptic spike changes the post-synaptic trace $Q$ according to  Eq. (2). If a pre-synaptic spike happens after a post-synaptic spike, weight will decrease by $\delta_{Wq}$, which is given by (3). If a post-synaptic spike occurs after a pre-synaptic spike, weight will increase by $\delta_{Wp}$ given by (4). 
\vspace{-0.8em}
\begin{equation}
P = P \times exp(\frac{t_{pre}\prime- t_{pre}}{ \tau_{_{LTP}} }) + ALTP 
\label{eq:1} 
\end{equation}
\vspace{-0.5em}
\begin{equation}
Q = Q \times exp(\frac{t_{post}\prime - t_{post}}{ \tau_{_{LTD}}  })+ ALTD 
\label{eq:1}
\end{equation}
\begin{equation}
\delta_{Wq} = aLTD \times Q \times exp(\frac{t_{post} - t_{pre}}{ \tau_{_{LTD}} } ), t_{post} < t_{pre}
\label{eq:1}
\end{equation}
\begin{equation}
\delta_{Wp} = aLTP \times P \times exp(\frac{t_{pre}-t_{post}}{ \tau_{_{LTP}}}), t_{pre} < t_{post} 
\label{eq:1}
\end{equation}
\vspace{-1em}

The STDP rule itself is not enough for learning. Both unsupervised \cite{unsupervised1} \cite{unsupervised2} \cite{unsupervised3} and supervised \cite{supervised1} \cite{supervised2} training algorithms based on the STDP rule have been proposed, which focus on computational aspect of the neuron model. These algorithms use Poisson based spike trains as input and include other bio-inspired mechanisms like winner-take-all \cite{WTA} and homoeostasis \cite{homo}.

Since artificial stimuli can be precisely applied to optically stimulated neurons, synchronous inputs are used for the proposed work. Considering a feedforward network, the synchronous inputs will lead to a problem. There will be no weight decrease for the network based on the basic STDP rule and all of the neurons will fire eventually. That is because when all of the input neurons fire at the same time, no post-synaptic neuron will fire before any pre-synaptic spike.

In order to solve this problem and make the living neural network easy to train, we propose a new supervised STDP training algorithm. Four basic operations in this algorithm are shown in Fig. 2 and discussed below. 

In this algorithm, stimuli can be applied to both input and output neurons artificially to generate a spike. In Fig. 2, external stimuli that generate a spike for input and output neurons are shown in pink and blue respectively.

Without artificial output stimuli, an output neuron can reach its action potential and fire in response to the network inputs, which is shown in yellow. Because the output neuron spikes after the input stimuli, weights between those in-out pairs will be naturally potentiated. This is referred to as the network's ``natural increase", with $t_{pre}-t_{post}=T2-T4$.  

Besides this natural increase, stimuli can be directly applied to the input and output neuron pairs to artificially change the weights. If an output stimulus is given at T3 and an input stimulus is given at T2, weight between this pair will be increased, which is referred to as ``artificial increase". If an output stimulus is given at T1, which is before an input stimulus, ``artificial decrease" will happen and weight will decrease.

In this work, these timing-based rules of applying external stimuli are the essential mechanisms to change synaptic weights. For the digit recognition task, groups of weights are increased or decreased to make the network converge. Some weights that are already reaching convergence need to be kept same. However, if the weight of a synapse is large enough to make the output neuron fire, the weight will increase naturally during the training process, which will move the network away from convergence. Therefore, each training step is separated into two phases, and the input stimuli corresponding to the input image are given once for each phase. During the first phase, if a weight that should be kept the same actually increased, a stimulus will be added to the corresponding output neurons at Tn+T0, which is before the input stimulus at Tn+T2 for the second phase to decrease the increased weight. Because of the refractory period, there will not be a natural increase at the second phase. The time interval between input and the stimuli (T2-T0) is adjusted with the time interval between natural output spike and input stimuli (T4-T2) at the first phase, so that the amount of decrease and increase matches. Through this approach, the weight can be kept roughly the same. This process is referred to as ``artificial hold".

\begin{figure*}[htb]
\begin{minipage}[b]{1.0\linewidth}
  \centering
  \vspace{-1.5em}
  \centerline{\includegraphics[width=18cm]{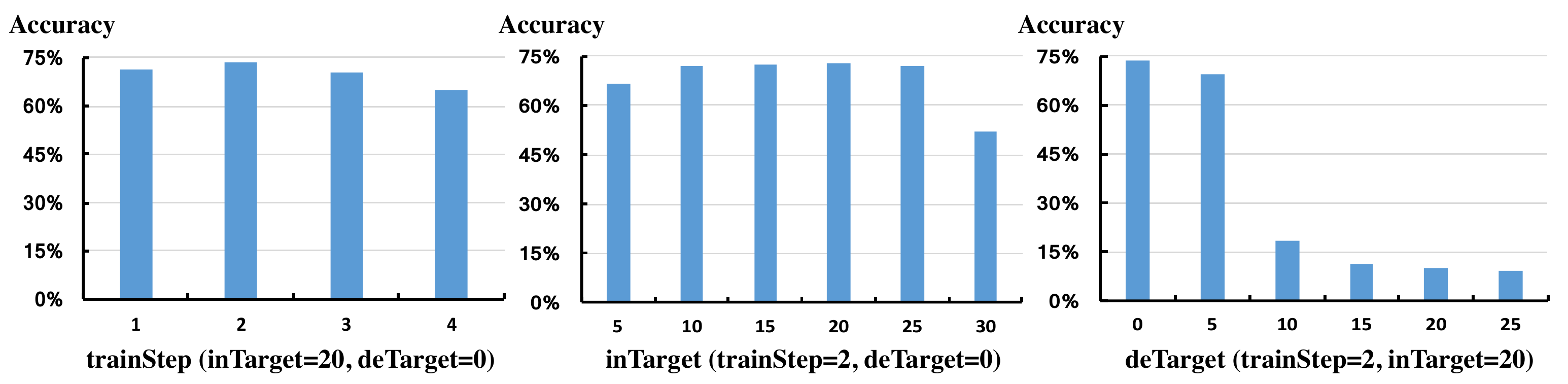}}
\end{minipage}
\vspace{-2.4em}
\caption{Sensitivity of the image recognition accuracy to different parameters.}
\vspace{-1.5em}
\label{fig:stdp}
\end{figure*}

The new STDP training algorithm is described in Algorithm 1. For each new image observed by the network, a prediction is made by applying the input stimuli to the network and checking the natural response of the outputs (Algorithm 1 lines 3-5). The index of the group that contain the largest number of spiking neurons is the predicted result (if two groups have the same number of spikes, the smaller group index is chosen). To train the network, the correct label (\textit{ID}) for the input image and the actual spike pattern for output neurons are compared to generate the control signals to select neurons into different lists that require external stimuli (lines 7-14). Based on the selected neurons, stimuli are applied to the network to update weights (lines 16-22). 

Three tunable parameters can be set for the training process. \textit{trainStep} is the number of training steps the network goes through for one image. A larger trainStep means higher effective learning rate. 
In order for the correct group to have more firing neurons than the other groups, \textit{inTarget} and \textit{deTarget} are targets of the number of firing neurons in each group. 
\textit{inTarget} represents the desired number of firing neurons to be observed in the output group that matches the correct label. 
\textit{deTarget} represents the desired number of firing neurons to be observed in the incorrect groups.
\textit{numSpike[id]} is the number of spiking neurons in group \textit{id}. 
When \textit{id} matches \textit{ID}, all spiking neurons in this group will be added to the \textit{holdList} to keep their weights since they respond correctly (line 8). If the number of firing neurons is less than \textit{inTarget},  \textit{inTarget-numSpike[id]} neurons will be randomly chosen among the non-spiking ones and added to the \textit{inList} (lines 9-11). 
For other output groups, if the number of firing neurons is more than \textit{deTarget}, \textit{numSpike[id]-deTarget} neurons will be randomly chosen among the spiking ones and added to the \textit{deList} (lines 12-14). 
After selecting \textit{inList}, \textit{deList}, and \textit{holdList}, corresponding stimuli are applied in time sequence shown in Fig. 2 for each training step (lines 16-22).

\vspace{-0.5em}
\begin{algorithm}
\caption{Supervised STDP Training}\label{euclid}
\begin{algorithmic}[1]
\State // {\bf{Tunable parameters}}: trainStep, inTarget, deTarget
\For {each image} 
\State Clear $inList$, $deList$ and $holdList$
\State Apply stimuli to input neurons based on pixel values
\State Record spike pattern for output neurons
	
	\For {each output group (id = 0 to 9)}

		\If {$id\! == \!ID$}
		\State Add all spiking neurons to $holdList$
			\If {$numSpike[id]\!\!<\!\!inTarget$}
			\State$x = inTarget-numSpike[id]$
			\State Add $x$ non-spiking neurons to $inList$
			\EndIf
		\EndIf
		
		\If {$id \ne ID$ {\bf{and}} $numSpike[id]\!\!>\!\!deTarget$}
		\State$y = numSpike[id]-deTarget$
		\State Add $y$ spiking neurons to $deList$
		\EndIf				
	\EndFor
	
	\For {each trainStep} 
		\State Apply stimuli to the $deList$ at T1	
		\State Apply stimuli to input neurons at T2
		\State Apply stimuli to the $inList$ at T3
		\State Apply stimuli to the $holdList$ at Tn+T0
		\State Apply stimuli to the $deList$ at Tn+T1
		\State Apply stimuli to input neurons at Tn+T2
		\State Apply stimuli to the $inList$ at Tn+T3	
	\EndFor	
\EndFor
\end{algorithmic}
\end{algorithm}
\vspace{-1.2em}

\vspace{-1em}
\section{Simulation}
\vspace{-1em}
\subsection{Parameters}
\vspace{-0.5em}
\label{sec:pagestyle}

Parameters used for the network are listed in Table 1. Timing parameters in Fig. 2 impact the learning rate for each \textit{trainStep}. Small time interval between pre- and post-synaptic neuron spikes (\textit{e.g.}, T3-T2) leads to greater weight changes. However, the optical stimuli cannot be spaced arbitrarily close to each other. In this work, a fixed 5 ms interval is used between pre- and post-synaptic spikes for artificial increase and decrease. The pre- and post-synaptic spike interval for the hold mechanism matches the natural increase interval, which is 10 ms according to empirical data. 
\vspace{-0.5em}
\begin{table}[!hbp]
\centering
\begin{tabular}{|c|c|c|}
\hline
\bf{Parameter} & \bf{Value} \\
\hline
$\tau_{_{LTP}}$  / $\tau_{_{LTD}}$  & 20ms / 20ms \\
ALTP / ALTD & 1 / -1  \\
aLTP / aLTD &  $6\times10^{-5}$  / $6.3\times10^{-5}$\\
max weight & 0.02 \\
refractory period& 25ms \\
T3-T2 / T2-T1 & 5ms / 5ms \\
T4-T2 / T2-T0 & 10ms / 10ms \\
Tn & 30ms \\
\hline
\end{tabular}
\vspace{-0.5em}
\caption{Simulation parameters, \cite{parameters}\cite{resting}\cite{ATDP}\cite{unsupervised1}.}
\vspace{-1.5em}
\end{table}

\vspace{-0.8em}
\subsection{Results and analysis}
\vspace{-0.5em}
Three tunable parameters for the network are: \textit{trainStep}, \textit{inTarget} and \textit{deTarget}. \textit{trainStep} is kept at 2, \textit{inTarget} is set at 20, and \textit{deTarget} is configured as 0 as the base line. A sensitivity study is done by tuning one parameter at a time. 

For this sensitivity study, 1000 images from the MNIST dataset are used for training, another 1000 images are used for testing. Training time for each trainStep is 60 ms. Simulation results are shown in Fig. 3. 

When \textit{trainStep} increases from 1 to 2, the prediction accuracy increases. However, when learning rate is too large (beyond 3 steps), the accuracy drops. This is because a larger effective learning rate may lead to fast convergence, but if it is too large, overshooting will happen when moving towards the global optimum point, which leads to oscillations and hurts the performance.

For \textit{inTarget}, better results are achieved in the middle range, which shows that, training nearly half of the neurons to fire can provide enough information while avoid divergence. For example, if two images have a large set of overlapping firing neurons in the inputs but have different labels, strengthening the connections between all of the firing inputs and the corresponding outputs for one image will likely lead to mis-prediction for the other image. The best performance is achieved at \textit{inTarget}=20.

Decreasing weights associated with all firing neurons in all incorrect groups (\textit{deTarget}=0) achieves the best performance. When \textit{deTarget} is larger than 10, performance drops dramatically. This is because inTarget is 20 for this set of results. The number of firing neurons in the incorrect group needs to be below 20 to make sure that the correct group has the greatest number of firing neurons. However, training 2 steps cannot guarantee both \textit{inTarget} and \textit{deTarget} are reached. The best accuracy for the sensitivity study is 72.7\%.

A larger dataset (10000 images from MNIST) is evaluated based on the best parameters: \textit{trainStep}=2, \textit{inTarget}=20 and \textit{deTarget}=0. The accuracy for this larger dataset is 74.7\%.

Compared to a single-layer fully connected ANN, which achieves 88\% \cite{onelayerANN} accuracy on MNIST dataset, the proposed supervised STDP-based SNN still has an accuracy gap. 
Unlike the ANN, where weights and inputs can be directly used in the prediction, most SNN models only rely on the output spikes other than the exact membrane potential to make a prediction. The loss of information leads to the accuracy drops. 
The only single-layer SNN work in our knowledge \cite{onelayerSNN} derives an extra mathematical function to extract more informations through the timing relationships for output spikes, which does not consider biological properties of neurons and synapses and hence is an unrealistic scheme for living neuron experiment.

For SNN works based on neuron science simulations, a three-layer design with supervised STDP achieved an accuracy of 75.93\% for 10 digit recognition task on a MNIST dataset \cite{supervised2}. That is similar to the proposed single-layer network in this paper. There are other SNN works that have better results on MNIST \cite{unsupervised1} \cite{unsupervised2} \cite{unsupervised3} \cite{supervised1} \cite{supervised3}. However, those networks have at least two layers and a larger number of neurons (e.g. 71,026 in \cite{supervised3}). 
Some works also preprocess the input images to achieve better accuracy \cite{supervised1}.
The major difference between the proposed work and prior works is that prior works are optimized for solid state computers. For the proposed supervised scheme, bioengineering constraints are considered. 
Input data are compressed and applied as synchronous spike trains, HH model are used instead of the integrate and fire model. The biological limitations on maintaining the synaptic weights lead to the design of two training phases and multiple training steps.

\vspace{-1em}
\section{Conclusion}
\label{sec:typestyle}
\makeatletter
\def\BState{\State\hskip-\ALG@thistlm}
\makeatother
\vspace{-1em}

To explore the possibility of using living neuron for machine learning tasks, a new supervised STDP training algorithm has been proposed and simulated on a fully-connected neural network based on the HH model. A 74.7\% accuracy is achieved on digit recognition task for the MNIST dataset. This result demonstrates the feasibility of using living neurons as computation elements for machine learning tasks.



\vfill\pagebreak

\begin{spacing}{0.83}
{
\small
\bibliographystyle{IEEEbib}
\bibliography{strings,refs}
}
\end{spacing} 

\end{document}